\documentclass[runningheads]{llncs}

\usepackage[utf8]{inputenc}
\usepackage[T2A,T1]{fontenc}
\usepackage[main=english,russian]{babel}
\usepackage{amsmath}
\usepackage{booktabs}
\usepackage{graphicx}
\usepackage{microtype}
\usepackage{tabularx}
\usepackage{url}
\usepackage{xurl}
\usepackage[hidelinks,breaklinks]{hyperref}

\hypersetup{
  pdftitle={Natural Language Code Retrieval for 1C:Enterprise: An Open Benchmark and Efficient Bi-Encoder},
  pdfauthor={Konstantin Chesnokov and Chingiz Mingazov},
  pdfkeywords={code retrieval, 1C:Enterprise, BSL, dense retrieval, benchmark, Matryoshka embeddings}
}

\newcommand{\bench}{\href{https://huggingface.co/datasets/PruhaNLP/1C-Ebench}{\nolinkurl{PruhaNLP/1C-Ebench}}}
\newcommand{\harness}{\href{https://github.com/PruhaNLP/1C-RB}{\nolinkurl{PruhaNLP/1C-RB}}}
\newcommand{\trainset}{\href{https://huggingface.co/datasets/PruhaNLP/1C-Code-Train}{\nolinkurl{PruhaNLP/1C-Code-Train}}}
\newcommand{\ourmodel}{\href{https://huggingface.co/PruhaNLP/USER2-1C-code}{\nolinkurl{PruhaNLP/USER2-1C-code}}}
\newcommand{\base}{\href{https://huggingface.co/deepvk/USER2-base}{\nolinkurl{deepvk/USER2-base}}}
\newcommand{\rumodernbert}{\href{https://huggingface.co/deepvk/RuModernBERT-base}{\nolinkurl{deepvk/RuModernBERT-base}}}
\newcommand{\embeddinggemma}{\href{https://huggingface.co/google/embeddinggemma-300m}{\nolinkurl{google/embeddinggemma-300m}}}
\newcommand{\bgemodel}{\href{https://huggingface.co/deepvk/USER-bge-m3}{\nolinkurl{deepvk/USER-bge-m3}}}
\newcommand{\granitemodel}{\href{https://huggingface.co/ibm-granite/granite-embedding-311m-multilingual-r2}{\nolinkurl{ibm-granite/granite-embedding-311m-multilingual-r2}}}
\newcommand{\harriermodel}{\href{https://huggingface.co/microsoft/harrier-oss-v1-270m}{\nolinkurl{microsoft/harrier-oss-v1-270m}}}
\newcommand{\efivemodel}{\href{https://huggingface.co/intfloat/multilingual-e5-base}{\nolinkurl{intfloat/multilingual-e5-base}}}
\newcommand{\sbertmodel}{\href{https://huggingface.co/ai-forever/sbert_large_nlu_ru}{\nolinkurl{ai-forever/sbert_large_nlu_ru}}}
\newcommand{\smallmodel}{\href{https://huggingface.co/deepvk/USER2-small}{\nolinkurl{deepvk/USER2-small}}}
\newcommand{\querymodel}{\href{https://huggingface.co/google/gemma-4-26B-A4B-it}{\nolinkurl{google/gemma-4-26B-A4B-it}}}
\providecommand{\Description}[1]{}
\newcommand{\ndcg}{nDCG@10}
\newcolumntype{Y}{>{\raggedright\arraybackslash}X}

\begin{document}

\title{Natural Language Code Retrieval for 1C:Enterprise: An Open Benchmark and Efficient Bi-Encoder}
\titlerunning{Natural Language Code Retrieval for 1C:Enterprise}

\author{Konstantin Chesnokov\inst{1}\orcidID{0009-0007-0162-9344}\textsuperscript{*} \and\nolinebreak
\mbox{Chingiz Mingazov\inst{2}\orcidID{0009-0006-1744-0207}}}
\authorrunning{K. Chesnokov and C. Mingazov}
\institute{Independent Researcher, Moscow, Russia\\
\email{konstphx@gmail.com}\quad\textsuperscript{*}Corresponding author
\and
Independent Researcher, Kazan, Russia}

\maketitle

\begin{abstract}
Natural language code retrieval is a rapidly evolving task in computer science. However, the 1C:Enterprise ecosystem combines Russian syntax with highly domain-specific terminology, for which open datasets and specialized models have been virtually non-existent. We present a comprehensive pipeline for 1C code retrieval: an open benchmark of 3,413 real-world, PII-scrubbed query--code pairs, a reproducible evaluation harness, and a specialized bi-encoder. To overcome scarce labeled data, we fine-tune on 784{,}057 synthetic triplets generated by \querymodel{} from public code repositories, using Matryoshka Representation Learning (MRL) and a privacy-aware tokenizer. Because the benchmark subsets differ in size, we report balanced-subset macro, query-weighted micro, and \texttt{forum}-only results. Our model reaches 0.5992 balanced macro nDCG@10, 0.5044 micro, and 0.4617 on \texttt{forum}, versus 0.4932 macro for the baseline architecture and 0.5404 for \embeddinggemma. Removing every benchmark example flagged by the conservative exact/13-gram overlap audit leaves 0.6011 balanced macro (0.5010 micro), indicating that detected train--benchmark overlap does not explain the headline result. MRL truncation to 256 dimensions preserves 99.9\% of retrieval quality while reducing dense-index storage and exact similarity arithmetic by a factor of three.

\end{abstract}

\section{Introduction}

Natural-language code retrieval ranks code fragments in response to a text query. For popular languages, sizeable datasets and trained retrievers exist; for 1C:Enterprise, open retrieval infrastructure has been scarce despite its role in Russian-speaking enterprise software.

The 1C domain differs from typical English-language code search. Queries are often Russian and may describe errors, platform objects, reports, registries, or accounting operations; relevant documents are BSL or 1C query fragments that mix Cyrillic keywords, Russian identifiers, SQL-like clauses, and business terminology. Without a domain benchmark, it is hard to tell whether general multilingual embeddings suffice or domain adaptation is needed.

We formulate the task as single-stage dense closed-set retrieval, where queries and documents are represented by learned continuous vectors and ranked by vector similarity. Given a corpus \(D=\{d_1,\ldots,d_n\}\) and query \(q\), the system ranks documents so that the labeled relevant fragment \(d^+\) ranks as high as possible. A bi-encoder embeds queries and documents independently:
\begin{equation}
\begin{aligned}
\mathbf{e}_q
&= \mathrm{Enc}\bigl(q;\;\texttt{search\_query}\bigr),
\\
\mathbf{e}_d
&= \mathrm{Enc}\bigl(d;\;\texttt{search\_document}\bigr),
\end{aligned}
\end{equation}
and scores them by cosine similarity (an inner product after L2 normalization):
\begin{equation}
\mathrm{score}(q,d)
=
\cos(\mathbf{e}_q,\mathbf{e}_d).
\end{equation}

This work establishes a starting point for 1C retrieval. We do not propose a new architecture or loss; we contribute a benchmark, evaluation protocol, training resource, contamination audit, and reference domain model.

The main contributions are:
\begin{enumerate}
\item \bench, an open benchmark for retrieving 1C/BSL code from Russian questions (3,413 pairs; subsets \texttt{forum} and \texttt{fastcode}).
\item \trainset, 784,057 synthetic triplets \((q,d^+,d^-)\) with PII scrubbing.
\item \harness, a reproducible evaluation harness for dense models and BM25Okapi.
\item A train--benchmark contamination audit (exact match and 13-gram); no exact duplicate pairs.
\item A reference baseline suite and \ourmodel, a domain-adapted bi-encoder with Matryoshka truncation that improves over \base{} and strong multilingual embeddings.
\end{enumerate}

\section{Related Work}

\subsection{Code Retrieval Benchmarks}

CodeSearchNet~\cite{codesearchnet} introduced a large corpus and challenge for semantic code search over six mainstream languages. CoSQA~\cite{cosqa} moved closer to real search with web queries. CodeXGLUE~\cite{codexglue} aggregated program-understanding tasks, including code search. CoIR~\cite{coir} covers text-to-code, code-to-code, and hybrid retrieval, while CodeRAG-Bench~\cite{coderagbench} evaluates retrieval for code generation. These resources cover widely used languages, but not 1C:Enterprise or Russian questions to BSL/1C query code.

\subsection{1C:Enterprise Evaluation Resources}

Open 1C evaluation has focused mainly on generation rather than retrieval. The 1C Code Bench leaderboard~\cite{onecodebench} evaluates BSL function generation with compilation and correctness metrics; PRISM/GenLab-1C develops executable BSL evaluation along similar lines~\cite{prismgenlab}. These resources are complementary to \bench: they ask whether a model can \emph{write} 1C code, whereas we ask whether an embedding model can \emph{find} a relevant fragment from a Russian query.

\subsection{Multilingual and Cross-lingual Retrieval Context}

Non-English software-engineering resources are increasingly needed as open-source collaboration becomes more multilingual~\cite{nonenglishoss}. \bench{} is not a classic cross-language information retrieval (CLIR) benchmark, because queries and much of the code vocabulary are Russian or Cyrillic-heavy rather than two separated natural languages; it still shares CLIR's alignment problem of matching intent in one form to artifacts in another~\cite{clirsurvey}. Downstream, domain code retrievers can help retrieval-augmented code generation~\cite{racg}; \bench{} can serve as a first-stage testbed for 1C. Generative retrieval and multilingual semantic compression~\cite{mgrcsc} are complementary future directions.

\subsection{Dense Retrieval, BM25, and Compact Embeddings}

Dense passage retrieval popularized scalable dual encoders~\cite{dpr}. BM25 remains a strong lexical baseline when exact API names or error strings matter~\cite{bm25}. Hard negatives from top-ranked candidates strengthen dense training~\cite{ance,rocketqa}; code-search work also studies false-negative risk~\cite{softinfonce,cocohanere}. Hybrid BM25+dense fusion via RRF~\cite{rrf} helps only when the lexical channel is complementary.

General multilingual embeddings such as \base, \bgemodel, \efivemodel, \embeddinggemma, and \granitemodel{} are useful baselines but are not optimized for Russian questions paired with 1C code. \ourmodel{} follows the \base{} sentence-transformers pipeline with Cached Multiple Negatives Ranking Loss and Matryoshka Representation Learning (MRL) for compact prefix embeddings---the first \(m\) dimensions of the full embedding---at inference~\cite{matryoshka}.

\subsection{Synthetic Queries and LLM-as-a-Judge}

When labeled query--code pairs are scarce, training often relies on synthetic queries, as in parts of CoIR. Quality is commonly screened with LLM-as-a-judge protocols~\cite{llmjudge}. Strong judges can agree with humans after bias mitigation on chat tasks, but this does not imply expert agreement in specialized domains; we therefore treat LLM-judge scores as an approximate signal, not ground truth.

\section{Benchmark and Evaluation Protocol}

\subsection{Resource Stack}

Table~\ref{tab:stack} summarizes the resource stack. The public benchmark, evaluation harness, and model are linked through their canonical Hugging Face and GitHub identifiers. They enable comparisons without access to private 1C systems. Provenance and upstream license metadata are preserved; we make no stronger claim about compatibility of all upstream licenses and exclude proprietary 1C platform distributions.

\begin{table}
\caption{Public resource stack.}
\label{tab:stack}
\centering
\begin{tabularx}{\textwidth}{lYl}
\toprule
Component & Content & Scale \\
\midrule
\bench & Test pairs: question \(\rightarrow\) code, subsets \texttt{forum}/\texttt{fastcode} & 3,413 pairs \\
\harness & Evaluation harness: dense, BM25, metrics & CLI/Python \\
\trainset & Triplets \((q,d^+,d^-)\) with PII scrubbing & 784,057 triplets \\
\ourmodel & Domain-adapted reference bi-encoder & 768d, MRL \\
\bottomrule
\end{tabularx}
\end{table}

\subsection{The \bench{} Benchmark}

\bench{} is built from real 1C questions and solutions rather than synthetic queries. The main source is the public Hugging Face dataset \texttt{arefaste/1C\_Forums} (19,041 records)~\cite{arefaste1cforums}. We extract Markdown code fences from \texttt{solution} and apply precision-first filters for BSL or 1C query fragments.

The benchmark-construction branch in Figure~\ref{fig:dataeval} summarizes the preparation pipeline: source records are normalized, candidate code is extracted, precision-first validity filters and exact deduplication are applied, and the remaining examples undergo privacy processing. In code, PII is replaced by placeholders such as \texttt{[EMAIL]}, \texttt{[PHONE]}, \texttt{[IP]}, \texttt{[PATH]}, \texttt{[PERSON]}, and \texttt{[REDACTED]}. Questions are not edited: residual PII triggers removal of the example. The scrubber is a rule-based, precision-oriented risk-reduction step over common surface forms, not a formal de-identification guarantee; we do not report PII detection precision/recall or a residual manual privacy audit.

The final benchmark has 3,413 examples in two subsets (Table~\ref{tab:subsets}). Each example has \texttt{id}, \texttt{question}, and \texttt{code}. Within a subset, the corpus is all \texttt{code} fields and relevance matches \texttt{id}. The query relevance judgments (qrels) assign one binary gold document to each query. Thus, the protocol is closed-set retrieval with a single gold; this can underestimate quality when an unannotated alternative ranks above gold.

\begin{table}
\caption{\bench{} subsets.}
\label{tab:subsets}
\centering
\begin{tabular}{@{}lrrlp{4.8cm}@{}}
\toprule
Subset & Queries & Share & Source & Characterization \\
\midrule
\texttt{forum} & 2{,}883 & 84.5\% & Forum discussions & Long questions, context, errors, inline code \\
\texttt{fastcode} & 530 & 15.5\% & Snippet catalog & Short queries, long ready-to-use snippets \\
\textbf{Total} & \textbf{3{,}413} & \textbf{100\%} & -- & Closed-set, single-gold qrels \\
\bottomrule
\end{tabular}
\end{table}

\subsection{Evaluation Harness}

\begin{figure}[htbp]
\centering
\includegraphics[width=\textwidth]{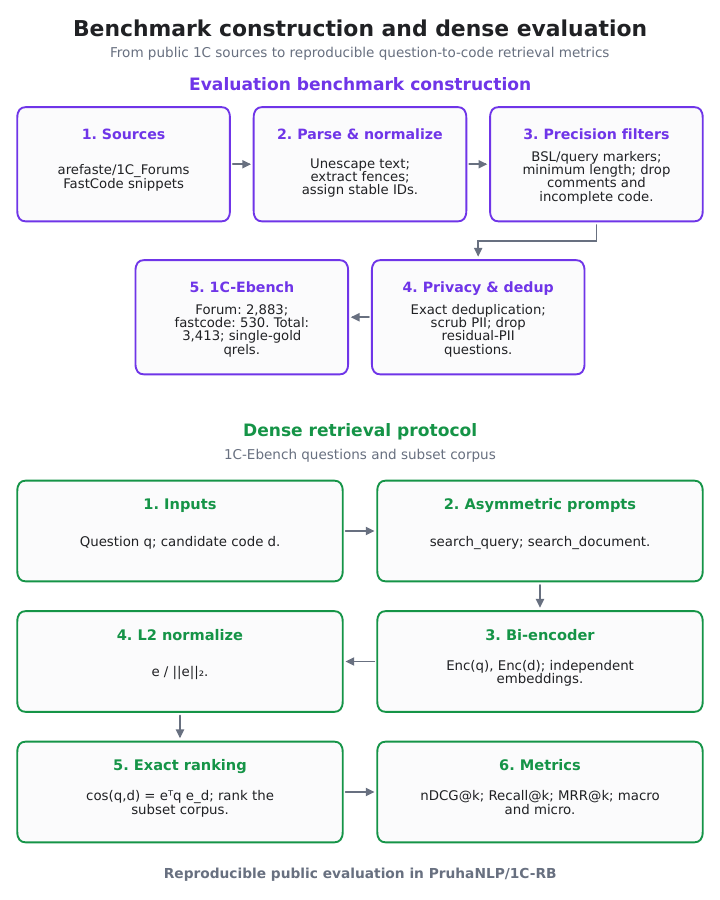}
\Description{Two-stage workflow diagram. The upper branch converts public 1C forum and FastCode sources through parsing, filtering, privacy scrubbing, and deduplication into the 3,413-pair 1C-Ebench benchmark. The lower branch embeds queries and candidate code with asymmetric prompts, applies L2 normalization and exact cosine ranking, and reports nDCG, recall, and MRR with macro and micro aggregation.}
\caption{Construction of the public \bench{} benchmark and the dense-retrieval evaluation pipeline implemented in \harness.}
\label{fig:dataeval}
\end{figure}

\harness{} loads \bench{} and builds \texttt{corpus}, \texttt{queries}, and \texttt{qrels}. The dense-retrieval branch in Figure~\ref{fig:dataeval} shows the complete evaluation pipeline: asymmetric query/document prompting, independent bi-encoder inference, L2 normalization, exact cosine ranking over the subset corpus, and computation of nDCG@\(k\), Recall@\(k\), and MRR@\(k\).

The headline metric is \ndcg{} for binary single-relevant qrels. Recall@\(k\) is 1 iff \(d^+\) is in the top \(k\); MRR@\(k\) is \(1/\mathrm{rank}(d^+)\) if that rank is \(\le k\), else 0. Let \(s_{\mathrm{forum}}\) and \(s_{\mathrm{fastcode}}\) denote the subset scores. The balanced-subset macro average assigns equal weight to the two retrieval regimes,
\(\mathrm{macro}=\tfrac{1}{2}(s_{\mathrm{forum}}+s_{\mathrm{fastcode}})\),
while micro-average weights by query volume
(\(n_{\mathrm{forum}}=2883\), \(n_{\mathrm{fastcode}}=530\)).
Thus, macro intentionally gives \texttt{fastcode} 50\% weight despite its 15.5\% query share; it estimates regime-balanced rather than random-query performance. We report query-weighted micro and per-subset scores alongside macro.

\subsection{Train--Benchmark Contamination Audit}

\trainset{} and \bench{} come from different sources: synthetic LLM queries over public GitHub 1C code versus real forum/FastCode questions. We still audit overlap with criteria from contamination studies. Dodge et al.~\cite{dodge2021c4} distinguish input-only, label-only, and input-and-label matches after normalization; Brown et al.~\cite{gpt3} use a conservative word \(N\)-gram dirty flag. Our protocol applies NFKC, lowercasing, whitespace collapsing, exact matching on code/question/pairs, and 13-gram dirty flags (Table~\ref{tab:contamination}).

\begin{table}
\caption{Train--benchmark overlap audit.}
\label{tab:contamination}
\centering
\begin{tabular}{lrrr}
\toprule
Metric & \texttt{forum} & \texttt{fastcode} & all \\
\midrule
Exact match: code & 0.10\% & 0.00\% & 0.09\% \\
Exact match: question & 0.03\% & 8.68\% & 1.38\% \\
Exact match: pair & 0.00\% & 0.00\% & \textbf{0.00\%} \\
13-gram overlap: code & 6.24\% & 16.23\% & 7.79\% \\
13-gram overlap: question & 0.03\% & 8.68\% & 1.38\% \\
Input-only & 0.03\% & 6.60\% & 1.05\% \\
Label-only & 6.24\% & 14.15\% & 7.47\% \\
Input-and-label & 0.00\% & 2.08\% & 0.32\% \\
Any 13-gram dirty flag & 6.28\% & 22.83\% & 8.85\% \\
\bottomrule
\end{tabular}
\end{table}

Under the strict exact-pair criterion, no train--benchmark duplicate pairs remain, and \texttt{fastcode} has no exact code matches. Its 22.83\% dirty rate decomposes into 14.15\% label-only, 6.60\% input-only, and 2.08\% input-and-label cases. All 46 exact-question matches in \texttt{fastcode} contain at most 12 words and are generic catalog-style titles, such as ``Copy an array'' or ``Path to a file.'' For texts shorter than 13 tokens, the conservative 13-gram criterion compares the entire text; consequently, the 8.68\% question 13-gram rate identifies the same short questions as exact matching rather than an additional disjoint set. The higher \texttt{fastcode} overlap therefore largely reflects short query templates and recurring BSL boilerplate.

\section{Domain Adaptation: \ourmodel}

\subsection{Model}

\ourmodel{} is a fine-tuned \base{} sentence-transformers bi-encoder with mean pooling and cosine scoring. Asymmetric prompts use different role-specific templates for queries and documents. Their names are \texttt{search\_query} and \texttt{search\_document} (Table~\ref{tab:model}).

We selected \base{} because it combines a strong Russian retrieval prior with practical fine-tuning. The 149M-parameter model, developed by DeepVK as a Universal Sentence Encoder for Russian, is based on \rumodernbert{} and supports contexts up to 8,192 tokens. Its published training pipeline comprises retrieval-oriented RetroMAE pretraining, weakly supervised English--Russian transfer, 50 million pairs mined from the Russian \texttt{cultura\_ru\_edu} corpus, and supervised tuning on 4.3 million examples that include multiple Russian retrieval datasets. Its moderate size and sentence-transformers interface make domain adaptation substantially more tractable than fine-tuning a large generative model.

\begin{table}
\caption{\ourmodel{} configuration.}
\label{tab:model}
\centering
\begin{tabularx}{\textwidth}{lY}
\toprule
Parameter & Value \\
\midrule
Base model & \base \\
Architecture & ModernBERT/RuModernBERT sentence-transformers encoder \\
Embedding dimension & 768 \\
Maximum sequence length & 8,192 \\
Pooling & Mean pooling \\
Similarity & Cosine over L2-normalized embeddings \\
MRL dimensions & 768, 512, 384, 256, 128, 64, 32 \\
\bottomrule
\end{tabularx}
\end{table}

For anonymized data we apply a privacy-aware tokenizer patch: \path{[PATH]/[PERSON]} map to \path{[unused0]/[unused1]}; email/phone/IP placeholders use special USER2 tokens; new-token embeddings are initialized from the mean of corresponding subtokens.

\subsection{Training Data}

\begin{figure}[!htbp]
\centering
\includegraphics[width=\textwidth]{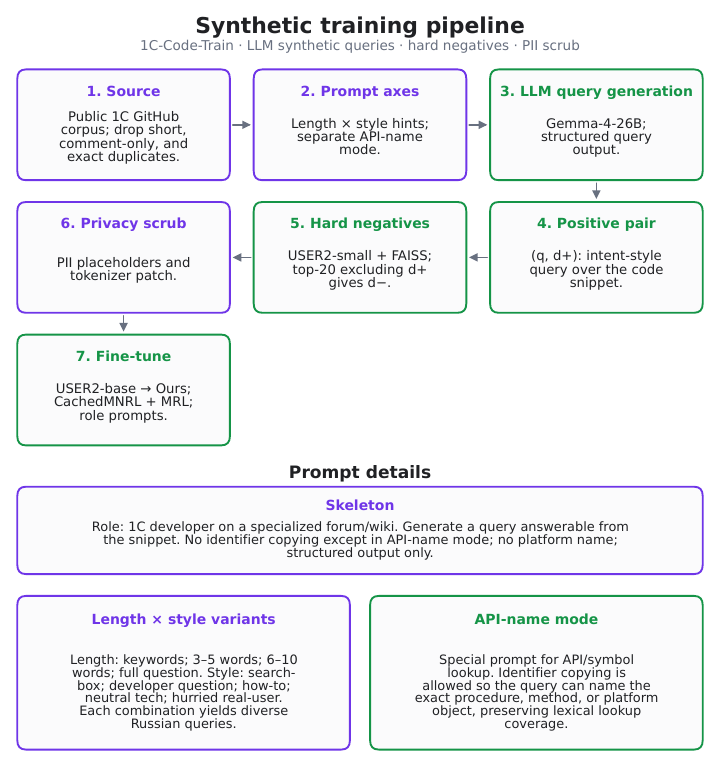}
\Description{Workflow diagram for synthetic training data. Public 1C code is filtered, query length and style prompts generate Russian queries, positive pairs and hard negatives are formed, privacy placeholders are applied, and USER2-base is fine-tuned with CachedMNRL and Matryoshka Representation Learning. A lower panel contrasts the default no-identifier prompt with an API-name mode that permits exact identifier copying.}
\caption{Construction of \trainset: from filtered public 1C code to synthetic queries, hard-negative triplets, PII scrubbing, and fine-tuning of \ourmodel.}
\label{fig:pipeline}
\end{figure}

With no open supervised 1C retrieval set available, we build \trainset{} as a synthetic closed loop from public code to contrastive triplets (Figure~\ref{fig:pipeline}). The design goal is coverage of both intent-style Russian questions and lexical API lookup, without turning the generator into a verbatim copier of identifiers.

Prompting is treated as a controlled sampling axis rather than a single fixed template: length and style hints diversify surface form, while a separate API-name mode deliberately allows identifier copying for symbol search. The remaining stages---hard-negative mining, privacy scrubbing, and USER2-style fine-tuning---are standard dense-retrieval engineering; Figure~\ref{fig:pipeline} makes their order explicit.

The source is the public Hugging Face corpus \path{leongl/1c_github} (3,038,637 lines)~\cite{leongl1cgithub}; after filtering short, comment-only, and duplicate fragments, 784{,}058 unique documents remain. For each document, \querymodel{} emits a Russian query as structured output. We mine hard negatives with \smallmodel{} using FAISS \texttt{IndexFlatIP}~\cite{faiss}. Specifically,
\(d^{-}=\arg\max_{d \in \mathrm{Top\text{-}20}(q)\setminus\{d^+\}}\cos(\mathbf{e}_q,\mathbf{e}_d)\).
Top hard negatives are standard~\cite{dpr,ance,rocketqa} but risk false negatives from near duplicates or alternative solutions~\cite{rocketqa,softinfonce}; we do not estimate the FN rate in \trainset. These two counts are not the same quantity: 784{,}058 is the filtered unique-document corpus, while 784{,}057 is the number of retained \((q,d^+,d^-)\) triplets after synthetic query generation (one filtered document does not yield a retained pair). PII anonymization changes 2,488 cells (\(\approx 0.106\%\)).

\subsection{Synthetic Query Quality}

We score a random sample of 300 triplets (seed 42) with \path{google/gemini-3.1-pro-preview} via OpenRouter~\cite{openrouter}, chosen because Gemini~3 Pro is strong on 1C Code Bench generation---an indirect, domain-relevant argument that is not the same as developer agreement on query quality. The judge rates relevance, naturalness, and clarity from 1 to 5 (Table~\ref{tab:judge}); we treat the scores as an approximate usability signal, not human validation.

\begin{table}
\caption{LLM-as-a-judge scores for 300 synthetic query--code pairs.}
\label{tab:judge}
\centering
\begin{tabular}{lrrr}
\toprule
Criterion & Mean & Median & Share \(\geq 4\) \\
\midrule
Relevance & 4.33 & 5 & 82.0\% \\
Naturalness & 4.17 & 5 & 76.3\% \\
Clarity & 4.37 & 5 & 82.0\% \\
\bottomrule
\end{tabular}
\end{table}

\subsection{Objective and Hyperparameters}

We follow the training recipe of \base. The objective is \(\mathcal{L}=\mathrm{MatryoshkaLoss}(\mathrm{CachedMNRL})\): InfoNCE~\cite{infonce} with one hard negative and in-batch negatives, summed over MRL prefixes \(\{768,512,384,256,128,64,32\}\). We keep this recipe for USER2 compatibility and Matryoshka truncation; the domain gain below is attributed to fine-tuning on \trainset, not to a novel loss (Table~\ref{tab:training}).

\begin{table}
\caption{Training hyperparameters.}
\label{tab:training}
\centering
\begin{tabular}{@{}ll@{}}
\toprule
Parameter & Value \\
\midrule
Training samples & 784{,}057 triplets \\
Batch size / MNRL mini-batch & 256 / 32 \\
Epochs / steps & 3 / 9{,}189 \\
Learning rate & \(2\times 10^{-5}\) \\
LR schedule & 5\% cosine warmup + decay \\
Max sequence length / precision & 8{,}192 / FP16 \\
Seed / hardware / runtime & 42 / A100 / \(\approx 7.5\)\,h \\
\bottomrule
\end{tabular}
\end{table}

\section{Experimental Setup}

\paragraph{Dense Models.}
We compare sentence-transformers-compatible models with fixed prompts (Table~\ref{tab:prompts}) under one \harness{} protocol, without exhaustive per-baseline tuning.

\begin{table}
\caption{Dense baselines and prompts. Empty means no prompt prefix.}
\label{tab:prompts}
\centering
\scriptsize
\begin{tabularx}{\textwidth}{Yll}
\toprule
Model & Query prompt & Document prompt \\
\midrule
\ourmodel & \texttt{search\_query} & \texttt{search\_document} \\
\base & \texttt{search\_query} & \texttt{search\_document} \\
\embeddinggemma & \texttt{query} & \texttt{document} \\
\bgemodel & empty & empty \\
\granitemodel & \texttt{query} & \texttt{document} \\
\harriermodel & \texttt{web\_search\_query} & empty \\
\efivemodel & \texttt{query:} & \texttt{passage:} \\
\sbertmodel & empty & empty \\
\bottomrule
\end{tabularx}
\end{table}

\paragraph{Lexical Baseline.}
BM25Okapi uses \(k_1=1.2\), \(b=0.75\), and identical Unicode tokenization \texttt{re.findall(r"\textbackslash w+", text.lower())} for queries and documents, with exact ranking over each subset corpus in \harness.

\paragraph{Hybrid RRF.}
We fuse BM25 and dense top-100 ranks with Reciprocal Rank Fusion (\(k_{\mathrm{RRF}}=60\)):
\(\mathrm{RRF}(q,d)=\sum_{m \in \{\mathrm{BM25},\,\mathrm{dense}\}} 1/(k_{\mathrm{RRF}}+\mathrm{rank}_{m}(q,d))\).

\section{Results}

\subsection{Main Leaderboard}

\begin{figure}
\centering
\includegraphics[width=.88\textwidth]{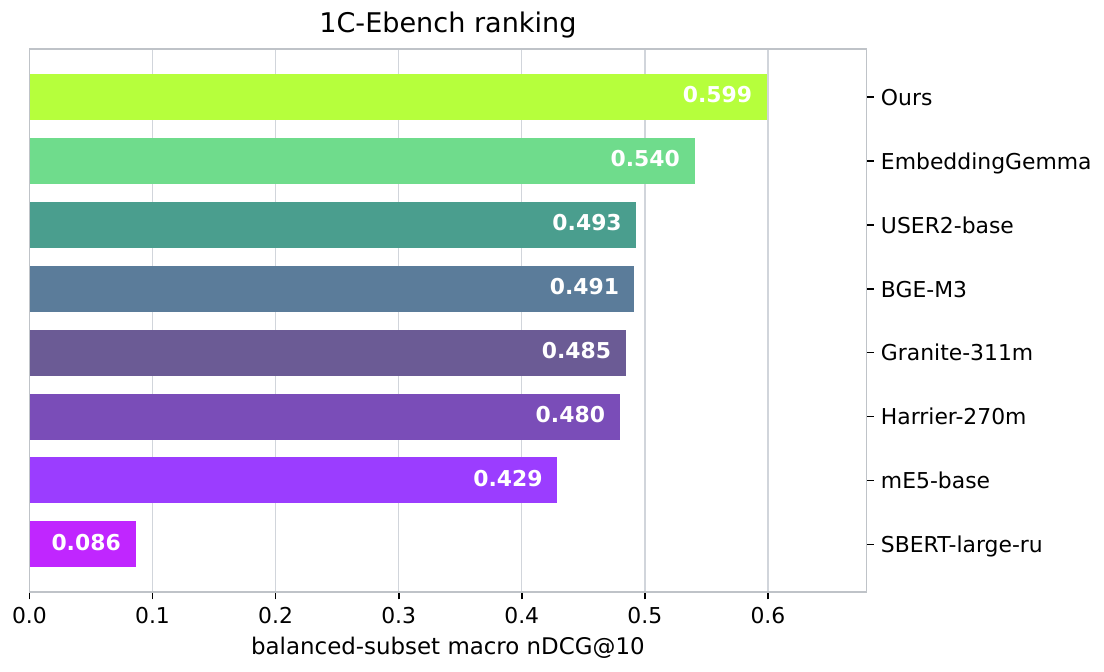}
\Description{Horizontal bar chart ranking eight dense models by balanced-subset macro nDCG at 10. PruhaNLP/USER2-1C-code ranks first at 0.599, followed by EmbeddingGemma at 0.540 and USER2-base at 0.493; sbert-large-nlu-ru ranks last at 0.086.}
\caption{\ndcg{} leaderboard on \bench{} (balanced-subset macro).}
\label{fig:leaderboard}
\end{figure}

Figure~\ref{fig:leaderboard} and Table~\ref{tab:leaderboard} show the main \ndcg{} leaderboard. \ourmodel{} obtains the best balanced-subset macro, 0.5992, while also ranking first under query-weighted micro (0.5044) and on \texttt{forum} alone (0.4617). Its macro gain is 0.106 over \base{} and 0.0588 over \embeddinggemma.

\begin{table}
\caption{\ndcg{} leaderboard on \bench.}
\label{tab:leaderboard}
\centering
\scriptsize
\begin{tabularx}{\textwidth}{rYrrrr}
\toprule
Rank & Model & \texttt{forum} & \texttt{fastcode} & macro & micro \\
\midrule
1 & \ourmodel & 0.4617 & 0.7366 & 0.5992 & 0.5044 \\
2 & \embeddinggemma & 0.3720 & 0.7080 & 0.5404 & 0.4242 \\
3 & \base & 0.3670 & 0.6190 & 0.4932 & 0.4061 \\
4 & \bgemodel & 0.3430 & 0.6390 & 0.4910 & 0.3890 \\
5 & \granitemodel & 0.3190 & 0.6510 & 0.4846 & 0.3706 \\
6 & \harriermodel & 0.3230 & 0.6360 & 0.4796 & 0.3716 \\
7 & \efivemodel & 0.3070 & 0.5510 & 0.4290 & 0.3449 \\
8 & \sbertmodel & 0.0940 & 0.0790 & 0.0862 & 0.0917 \\
-- & BM25Okapi & 0.2865 & 0.3334 & 0.3099 & 0.2938 \\
-- & RRF(BM25 + \ourmodel) & 0.4000 & 0.4600 & 0.4300 & 0.4093 \\
-- & RRF(BM25 + \base) & 0.3620 & 0.4195 & 0.3908 & 0.3709 \\
\bottomrule
\end{tabularx}
\end{table}

BM25Okapi lags strong dense models, suggesting lexical overlap is insufficient for many Russian-to-BSL queries. Untuned RRF hurts both dense systems, reducing balanced macro \ndcg{} from 0.5992 to 0.4300 for \ourmodel{} and from 0.4932 to 0.3908 for \base{}, with the largest drop on \texttt{fastcode}. A likely cause is weak lexical complementarity under naive \texttt{\textbackslash w+} tokenization, which poorly isolates BSL identifiers and error strings. These results are for the tested untuned RRF setup and do not rule out stronger hybrids.

Table~\ref{tab:fullmetrics} gives full @10 metrics for \ourmodel{} and BM25. \texttt{forum} is harder: longer contextual questions and single-gold qrels can penalize unannotated alternatives.

\begin{table}
\caption{Full @10 metrics for the reference dense model and BM25.}
\label{tab:fullmetrics}
\centering
\begin{tabular}{llrrr}
\toprule
System & Subset & nDCG@10 & Recall@10 & MRR@10 \\
\midrule
\ourmodel & \texttt{forum} & 0.4617 & 0.6008 & 0.4178 \\
\ourmodel & \texttt{fastcode} & 0.7366 & 0.9208 & 0.6774 \\
BM25Okapi & \texttt{forum} & 0.2865 & 0.3479 & 0.2670 \\
BM25Okapi & \texttt{fastcode} & 0.3334 & 0.4189 & 0.3067 \\
\bottomrule
\end{tabular}
\end{table}

\subsection{Controlled Comparisons}

\subsubsection{Domain Adaptation.}
Comparing \base{} and \ourmodel{} under the same architecture and prompts yields +0.106 balanced macro \ndcg. Paired bootstrap over queries (10,000 resamples, stratified by subset) gives 95\% CI \([0.087, 0.125]\) and one-sided \(P(\Delta\leq 0)<0.001\) (Table~\ref{tab:bootstrap}). Against \embeddinggemma, the balanced macro difference is +0.0588, with 95\% CI \([0.042, 0.075]\) and one-sided \(P(\Delta\leq 0)<0.001\). The same table shows untuned RRF vs.\ dense-only: significantly worse for \ourmodel{} on both subsets and for \base{} on \texttt{fastcode}.

\begin{table}
\caption{Paired-bootstrap \ndcg{} comparisons.}
\label{tab:bootstrap}
\centering
\small
\begin{tabularx}{\textwidth}{@{}Y>{\centering\arraybackslash}p{0.10\textwidth}>{\raggedright\arraybackslash}p{0.35\textwidth}@{}}
\toprule
Comparison & \(\Delta\) & 95\% CI / \(P\) \\
\midrule
Macro: \href{https://huggingface.co/PruhaNLP/USER2-1C-code}{\texttt{Ours}} vs.\ \href{https://huggingface.co/deepvk/USER2-base}{\texttt{USER2-base}}
  & $+0.106$ & $[0.087,\,0.125]$, \(P<0.001\) \\
Macro: \href{https://huggingface.co/PruhaNLP/USER2-1C-code}{\texttt{Ours}} vs.\ \href{https://huggingface.co/google/embeddinggemma-300m}{\texttt{EmbeddingGemma}}
  & $+0.0588$ & $[0.042,\,0.075]$, \(P<0.001\) \\
RRF vs.\ dense (\href{https://huggingface.co/PruhaNLP/USER2-1C-code}{\texttt{Ours}}), \texttt{forum}
  & $-0.062$ & $[-0.075,\,-0.049]$, \(P<0.05\) \\
RRF vs.\ dense (\href{https://huggingface.co/PruhaNLP/USER2-1C-code}{\texttt{Ours}}), \texttt{fastcode}
  & $-0.277$ & $[-0.319,\,-0.234]$, \(P<0.05\) \\
RRF vs.\ dense (\href{https://huggingface.co/deepvk/USER2-base}{\texttt{Base}}), \texttt{forum}
  & $-0.005$ & $[-0.016,\,0.006]$, n.s. \\
RRF vs.\ dense (\href{https://huggingface.co/deepvk/USER2-base}{\texttt{Base}}), \texttt{fastcode}
  & $-0.200$ & $[-0.237,\,-0.164]$, \(P<0.05\) \\
\bottomrule
\end{tabularx}
\end{table}

\subsubsection{Contamination Sensitivity.}
We recompute \ndcg{} after removing every example flagged by the conservative audit through an exact or 13-gram overlap in either the question or code. On the resulting clean subsets, \ourmodel{} scores 0.4653 on \texttt{forum} (\(n=2702\)) and 0.7369 on \texttt{fastcode} (\(n=409\)), giving 0.6011 balanced macro and 0.5010 query-weighted micro. The original \texttt{fastcode} score is 0.7366, compared with 0.7356 on flagged examples, 0.7359 after excluding exact-question matches, and 0.7357 after excluding input-and-label cases. Thus, the high \texttt{fastcode} score is stable under all tested exclusions. For comparison, \embeddinggemma{} decreases from 0.7084 to 0.6903 on clean-only \texttt{fastcode}; the clean-only margin of \ourmodel{} is therefore larger, not smaller.

\subsection{Qualitative Examples}

Fine-tuning helps when a query uses 1C platform vocabulary whose gold answer is a specific metadata or UI API rather than a shared lexical token. It can still fail on out-of-domain COM utilities and on short underspecified titles, where general multilingual encoders recover via surface cues. Table~\ref{tab:qualitative} shows two wins (\ourmodel{} at rank~1; every dense baseline in Table~\ref{tab:prompts} and BM25 outside the top~10) and two failures.

\begin{table}
\caption{Qualitative wins and failures (original Russian query/code). Gold rank: 1~=~best; \(>\)10~=~outside top-10.}
\label{tab:qualitative}
\centering
\footnotesize
\renewcommand{\arraystretch}{1.08}
\shorthandoff{"}%
\begin{tabularx}{\textwidth}{@{}>{\raggedright\arraybackslash}p{0.25\textwidth}Y@{}}
\toprule
Type / ranks & Query, gold, interpretation \\
\midrule
Win (\texttt{forum})
\newline Ours~1; others \(>\)10
  & \textbf{Q:} \foreignlanguage{russian}{Как обойти ограничения ФО на реквизит в динамическом списке, не трогая константу ФО?}
    \newline\textbf{Gold:} \foreignlanguage{russian}{ОтключенныеПоля = Новый Массив;} \ldots
    \newline\textit{FO\(\times\)dynamic-list pattern; little lexical overlap with gold.} \\
\addlinespace[5pt]
Win (\texttt{fastcode})
\newline Ours~1; others \(>\)10
  & \textbf{Q:} \foreignlanguage{russian}{Описание строки в таблице значений}
    \newline\textbf{Gold:} \foreignlanguage{russian}{ст90 = новый ОписаниеТипов("строка",\relax,Новый КвалификаторыСтроки(90));}
    \newline\textit{Colloquial} \foreignlanguage{russian}{«описание строки»} \textit{$\to$ type API; lexical baselines miss it.} \\
\addlinespace[5pt]
Fail (\texttt{fastcode})
\newline Ours \(>\)10; base~1
  & \textbf{Q:} \foreignlanguage{russian}{Произношение текста голосом}
    \newline\textbf{Gold:} \foreignlanguage{russian}{Voice = Новый COMObject("SAPI.SpVoice"); Voice.Speak("Привет!");}
    \newline\textit{Rare COM API; general models recover \texttt{Speak}/\texttt{SAPI}.} \\
\addlinespace[5pt]
Fail (\texttt{fastcode})
\newline Ours~32; base~1
  & \textbf{Q:} \foreignlanguage{russian}{Сжатие длинной Строки}
    \newline\textbf{Gold:} \foreignlanguage{russian}{СжатиеДанных = Новый СжатиеДанных(9);} \ldots Base64.
    \newline\textit{Short generic title; surface} \foreignlanguage{russian}{«сжатие»+«строка»} \textit{wins.} \\
\bottomrule
\end{tabularx}
\shorthandon{"}%
\end{table}

\subsection{Matryoshka Truncation}

\begin{figure}
\centering
\includegraphics[width=.80\textwidth]{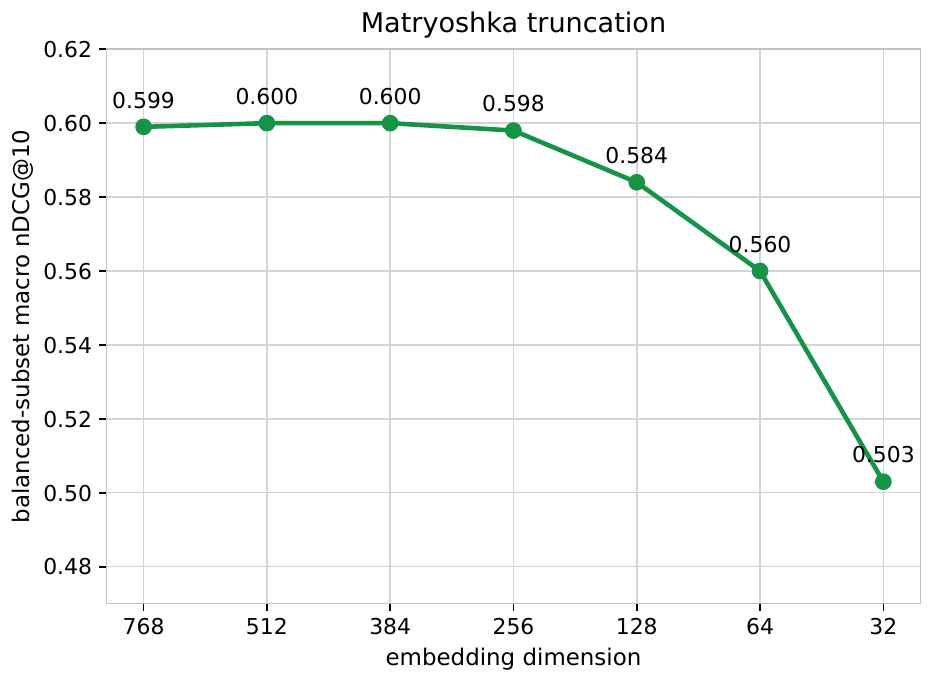}
\Description{Line chart of balanced-subset macro nDCG at 10 versus embedding dimension from 768 to 32. The score remains nearly flat through 256 dimensions and then declines from 0.598 at 256 dimensions to 0.503 at 32 dimensions.}
\caption{Matryoshka truncation: balanced-subset macro \ndcg{} by embedding dimension.}
\label{fig:mrl}
\end{figure}

MRL truncates a trained embedding to a prefix
\(\mathbf{e}^{(m)}=\mathbf{e}_{1:m}\)
and re-normalizes. Figure~\ref{fig:mrl} and Table~\ref{tab:mrl} show that 256-d embeddings retain 99.9\% of 768-d \ndcg{} at one third the dense-index size. Because cosine similarity between L2-normalized embeddings is a \(d\)-dimensional dot product, reducing \(d\) from 768 to 256 also cuts the per-comparison multiply--add count by a factor of three, yielding a corresponding theoretical \(3\times\) speedup in exact similarity scoring; end-to-end retrieval latency is not measured here.

\begin{table}
\caption{MRL truncation on \bench.}
\label{tab:mrl}
\centering
\begin{tabular}{rrrr}
\toprule
Dimension \(m\) & macro \ndcg{} & Retention & Relative index size \\
\midrule
768 & 0.599 & 100.0\% & \(1.00\times\) \\
512 & 0.600 & 100.0\% & \(0.67\times\) \\
384 & 0.600 & 100.0\% & \(0.50\times\) \\
256 & 0.598 & 99.9\% & \(0.33\times\) \\
128 & 0.584 & 97.5\% & \(0.17\times\) \\
64 & 0.560 & 93.5\% & \(0.08\times\) \\
32 & 0.503 & 83.9\% & \(0.04\times\) \\
\bottomrule
\end{tabular}
\end{table}

\section{Limitations and Future Work}

\subsubsection{Single-Gold Qrels.}
\bench{} has exactly one gold document per query. On forum data this can underestimate quality when alternatives exist. In a manual check of 50 queries (25 per subset, seed 42), top-10 mined neighbors contained multiple relevant fragments in 5 cases (10\%), mostly near duplicates or equivalent recipes. Incomplete judgments are well known in IR (e.g., BEIR~\cite{beir}) and also affect hard-negative mining; we neither estimate FN rate nor denoise in this version.

\subsubsection{Synthetic Data and Judge Validation.}
Synthetic query quality is screened with an LLM judge on 300 examples, without human--AI agreement in the specialized Russian-to-BSL domain. Scores are a useful signal but not expert annotation.

\subsubsection{Privacy and Licensing.}
PII scrubbing is rule-based over common surface patterns without measured precision/recall or a residual audit. License metadata and provenance are documented, but we do not resolve every redistribution implication of heterogeneous upstream sources.

\subsubsection{Experimental Coverage.}
We do not evaluate rerankers, multi-seed variance, hard-vs-random negatives, tokenizer-patch ablations, ANN indexes, or latency at larger scale. The train--benchmark domain gap remains, and \(n\)-gram contamination misses semantic paraphrases. Future work includes multi-gold qrels, human query judging, FN denoising, tuned hybrids/rerankers, and targeted ablations.

\section{Conclusion}

We presented an open stack for retrieving 1C:Enterprise code from Russian queries: \bench, \harness, \trainset, and reference bi-encoder \ourmodel. The resources provide a reproducible basis for a domain largely absent from open code-search benchmarks.

Domain adaptation matters: \ourmodel{} reaches 0.5992 balanced macro \ndcg{}, 0.5044 query-weighted micro, and 0.4617 on \texttt{forum}, beating \base{} by 0.106 and \embeddinggemma{} by 0.0588 in balanced macro under paired bootstrap. Removing every exact/13-gram-flagged example leaves 0.6011 balanced macro (0.5010 micro), while clean-only \texttt{fastcode} remains 0.7369; detected overlap therefore does not explain the headline result. Dense retrieval also substantially beats BM25Okapi, while naive RRF helps neither \ourmodel{} nor \base{} in the tested setup. Matryoshka truncation to 256-d preserves 99.9\% quality at roughly one third storage and similarity arithmetic.

\begin{credits}
\subsubsection{\discintname}
The authors have no competing interests to declare that are relevant to the content of this article.
\end{credits}

\bibliographystyle{splncs04}
\bibliography{references}

@article{codesearchnet,
  author       = {Husain, Hamel and Wu, Ho-Hsiang and Gazit, Tiferet and Allamanis, Miltiadis and Brockschmidt, Marc},
  title        = {{CodeSearchNet} Challenge: Evaluating the State of Semantic Code Search},
  journal      = {CoRR},
  volume       = {abs/1909.09436},
  year         = {2019},
  url          = {https://arxiv.org/abs/1909.09436}
}

@inproceedings{cosqa,
  author       = {Huang, Junjie and Tang, Duyu and Shou, Linjun and Gong, Ming and Xu, Ke and Jiang, Daxin and others},
  title        = {{CoSQA}: 20,000+ Web Queries for Code Search and Question Answering},
  booktitle    = {Proceedings of the 59th Annual Meeting of the Association for Computational Linguistics and the 11th International Joint Conference on Natural Language Processing (Volume 1: Long Papers)},
  pages        = {5690--5700},
  year         = {2021},
  doi          = {10.18653/v1/2021.acl-long.442}
}

@article{codexglue,
  author       = {Lu, Shuai and Guo, Daya and Ren, Shuo and Huang, Junjie and Svyatkovskiy, Alexey and Blanco, Ambrosio and others},
  title        = {{CodeXGLUE}: A Machine Learning Benchmark Dataset for Code Understanding and Generation},
  journal      = {CoRR},
  volume       = {abs/2102.04664},
  year         = {2021},
  url          = {https://arxiv.org/abs/2102.04664}
}

@inproceedings{coir,
  author       = {Li, Xiangyang and Dong, Kuicai and Lee, Yi Quan and Xia, Wei and Zhang, Hao and Dai, Xinyi and others},
  title        = {{CoIR}: A Comprehensive Benchmark for Code Information Retrieval Models},
  booktitle    = {Proceedings of the 63rd Annual Meeting of the Association for Computational Linguistics (Volume 1: Long Papers)},
  pages        = {22074--22091},
  year         = {2025},
  doi          = {10.18653/v1/2025.acl-long.1072}
}

@inproceedings{coderagbench,
  author       = {Wang, Zora Zhiruo and Asai, Akari and Yu, Xinyan Velocity and Xu, Frank F. and Xie, Yiqing and Neubig, Graham and others},
  title        = {{CodeRAG-Bench}: Can Retrieval Augment Code Generation?},
  booktitle    = {Findings of the Association for Computational Linguistics: NAACL 2025},
  pages        = {3199--3214},
  year         = {2025},
  doi          = {10.18653/v1/2025.findings-naacl.176}
}

@misc{onecodebench,
  author       = {{GigaCode R\&D} and {Sber AI}},
  title        = {{1C Code Bench}: A Benchmark for Evaluating the Ability of {LLMs} to Write {1C} Code},
  year         = {2026},
  howpublished = {Habr, \url{https://habr.com/ru/companies/sberbank/articles/1040114/}},
  note         = {In Russian. Accessed: 2026-08-18}
}

@misc{prismgenlab,
  author       = {{GenLab-1C}},
  title        = {{PRISM}: Executable {BSL} Code Generation Benchmark for {1C:Enterprise}},
  year         = {2026},
  howpublished = {\url{https://github.com/genlab-1c/prism}},
  note         = {Accessed: 2026-07-12}
}

@misc{arefaste1cforums,
  author       = {arefaste},
  title        = {{Arefaste}: Parsed Dataset for 1C from Popular Forums},
  year         = {2025},
  publisher    = {Hugging Face},
  howpublished = {\url{https://huggingface.co/datasets/arefaste/1C_Forums}},
  note         = {Accessed: 2026-07-12}
}

@misc{leongl1cgithub,
  author       = {leongl},
  title        = {{1c\_github}: 1C Code Corpus from {GitHub}},
  year         = {2024},
  publisher    = {Hugging Face},
  howpublished = {\url{https://huggingface.co/datasets/leongl/1c_github}},
  note         = {Accessed: 2026-07-12}
}

@inproceedings{dpr,
  author       = {Karpukhin, Vladimir and Oguz, Barlas and Min, Sewon and Lewis, Patrick and Wu, Ledell and Edunov, Sergey and others},
  title        = {Dense Passage Retrieval for Open-Domain Question Answering},
  booktitle    = {Proceedings of the 2020 Conference on Empirical Methods in Natural Language Processing (EMNLP)},
  pages        = {6769--6781},
  year         = {2020},
  doi          = {10.18653/v1/2020.emnlp-main.550}
}

@article{bm25,
  author       = {Robertson, Stephen and Zaragoza, Hugo},
  title        = {The Probabilistic Relevance Framework: {BM25} and Beyond},
  journal      = {Foundations and Trends in Information Retrieval},
  volume       = {3},
  number       = {4},
  pages        = {333--389},
  year         = {2009},
  doi          = {10.1561/1500000019}
}

@inproceedings{ance,
  author       = {Xiong, Lee and Xiong, Chenyan and Li, Ye and Tang, Kwok-Fung and Liu, Jialin and Bennett, Paul and others},
  title        = {Approximate Nearest Neighbor Negative Contrastive Learning for Dense Text Retrieval},
  booktitle    = {International Conference on Learning Representations},
  year         = {2021},
  url          = {https://openreview.net/forum?id=zeFrfgyZln}
}

@inproceedings{rocketqa,
  author       = {Qu, Yingqi and Ding, Yuchen and Liu, Jing and Liu, Kai and Ren, Ruiyang and Zhao, Wayne Xin and others},
  title        = {{RocketQA}: An Optimized Training Approach to Dense Passage Retrieval for Open-Domain Question Answering},
  booktitle    = {Proceedings of the 2021 Conference of the North American Chapter of the Association for Computational Linguistics: Human Language Technologies},
  pages        = {5835--5847},
  year         = {2021},
  doi          = {10.18653/v1/2021.naacl-main.466}
}

@inproceedings{softinfonce,
  author       = {Li, Haochen and Zhou, Xin and Tuan, Luu Anh and Miao, Chunyan},
  title        = {Rethinking Negative Pairs in Code Search},
  booktitle    = {Proceedings of the 2023 Conference on Empirical Methods in Natural Language Processing},
  pages        = {12760--12774},
  year         = {2023},
  doi          = {10.18653/v1/2023.emnlp-main.786}
}

@article{cocohanere,
  author       = {Fan, Ye and Li, Chuanyi and Ge, Jidong and Huang, LiGuo and Luo, Bin},
  title        = {Effective Hard Negative Mining for Contrastive Learning-Based Code Search},
  journal      = {{ACM} Transactions on Software Engineering and Methodology},
  volume       = {34},
  number       = {3},
  pages        = {1--35},
  year         = {2025},
  doi          = {10.1145/3695994}
}

@inproceedings{rrf,
  author       = {Cormack, Gordon V. and Clarke, Charles L. A. and Buettcher, Stefan},
  title        = {Reciprocal Rank Fusion Outperforms Condorcet and Individual Rank Learning Methods},
  booktitle    = {Proceedings of the 32nd International ACM SIGIR Conference on Research and Development in Information Retrieval},
  pages        = {758--759},
  year         = {2009},
  doi          = {10.1145/1571941.1572114}
}

@inproceedings{matryoshka,
  author       = {Kusupati, Aditya and Bhatt, Gantavya and Rege, Aniket and Wallingford, Matthew and Sinha, Aditya and Ramanujan, Vivek and others},
  title        = {Matryoshka Representation Learning},
  booktitle    = {Advances in Neural Information Processing Systems},
  volume       = {35},
  pages        = {30233--30249},
  year         = {2022},
  url          = {https://arxiv.org/abs/2205.13147}
}

@inproceedings{llmjudge,
  author       = {Zheng, Lianmin and Chiang, Wei-Lin and Sheng, Ying and Zhuang, Siyuan and Wu, Zhanghao and Zhuang, Yonghao and others},
  title        = {Judging {LLM}-as-a-Judge with {MT-Bench} and Chatbot Arena},
  booktitle    = {Advances in Neural Information Processing Systems},
  volume       = {36},
  pages        = {46595--46623},
  year         = {2023},
  url          = {https://arxiv.org/abs/2306.05685}
}

@inproceedings{dodge2021c4,
  author       = {Dodge, Jesse and Sap, Maarten and Marasovi{\'c}, Ana and Agnew, William and Ilharco, Gabriel and Groeneveld, Dirk and others},
  title        = {Documenting Large Webtext Corpora: A Case Study on the Colossal Clean Crawled Corpus},
  booktitle    = {Proceedings of the 2021 Conference on Empirical Methods in Natural Language Processing},
  pages        = {1286--1305},
  year         = {2021},
  doi          = {10.18653/v1/2021.emnlp-main.98}
}

@inproceedings{gpt3,
  author       = {Brown, Tom B. and Mann, Benjamin and Ryder, Nick and Subbiah, Melanie and Kaplan, Jared and Dhariwal, Prafulla and others},
  title        = {Language Models are Few-Shot Learners},
  booktitle    = {Advances in Neural Information Processing Systems},
  volume       = {33},
  pages        = {1877--1901},
  year         = {2020},
  url          = {https://arxiv.org/abs/2005.14165}
}

@inproceedings{beir,
  author       = {Thakur, Nandan and Reimers, Nils and R{\"u}ckl{\'e}, Andreas and Srivastava, Abhishek and Gurevych, Iryna},
  title        = {{BEIR}: A Heterogeneous Benchmark for Zero-shot Evaluation of Information Retrieval Models},
  booktitle    = {Proceedings of the Neural Information Processing Systems Track on Datasets and Benchmarks},
  year         = {2021},
  url          = {https://arxiv.org/abs/2104.08663}
}

@article{nonenglishoss,
  author       = {Bhuiyan, Masudul Hasan Masud and Kumar, Manish Kumar Bala and Staicu, Cristian-Alexandru},
  title        = {``Write in English, Nobody Understands Your Language Here'': A Study of Non-English Trends in Open-Source Repositories},
  journal      = {CoRR},
  volume       = {abs/2602.19446},
  year         = {2026},
  url          = {https://arxiv.org/abs/2602.19446}
}

@article{clirsurvey,
  author       = {Goworek, Roksana and Macmillan-Scott, Olivia and {\"O}zyi{\u{g}}it, Eda B.},
  title        = {Bridging Language Gaps: Advances in Cross-Lingual Information Retrieval with Multilingual LLMs},
  journal      = {CoRR},
  volume       = {abs/2510.00908},
  year         = {2025},
  url          = {https://arxiv.org/abs/2510.00908}
}

@article{mgrcsc,
  author       = {Huang, Yuxin and Wu, Simeng and Song, Ran and Xiang, Yan and Xian, Yantuan and Gao, Shengxiang and others},
  title        = {Multilingual Generative Retrieval via Cross-lingual Semantic Compression},
  journal      = {CoRR},
  volume       = {abs/2510.07812},
  year         = {2025},
  url          = {https://arxiv.org/abs/2510.07812}
}

@inproceedings{racg,
  author       = {Zhu, Qiming and Cao, Jialun and Chen, Xuanang and Zhang, Weili and Lu, Yaojie and Lin, Hongyu and others},
  title        = {Across Programming Language Silos: A Study on Cross-Lingual Retrieval-Augmented Code Generation},
  booktitle    = {Findings of the Association for Computational Linguistics: ACL 2026},
  pages        = {24283--24296},
  year         = {2026},
  doi          = {10.18653/v1/2026.findings-acl.1216}
}

@article{infonce,
  author       = {van den Oord, A{\"a}ron and Li, Yazhe and Vinyals, Oriol},
  title        = {Representation Learning with Contrastive Predictive Coding},
  journal      = {CoRR},
  volume       = {abs/1807.03748},
  year         = {2018},
  doi          = {10.48550/arXiv.1807.03748}
}

@article{faiss,
  author       = {Johnson, Jeff and Douze, Matthijs and J{\'e}gou, Herv{\'e}},
  title        = {Billion-Scale Similarity Search with {GPUs}},
  journal      = {{IEEE} Transactions on Big Data},
  volume       = {7},
  number       = {3},
  pages        = {535--547},
  year         = {2021},
  doi          = {10.1109/TBDATA.2019.2921572}
}

@misc{openrouter,
  author       = {{OpenRouter}},
  title        = {{OpenRouter} Documentation},
  year         = {2026},
  howpublished = {\url{https://openrouter.ai/docs}},
  note         = {Accessed: 2026-08-18}
}

\end{document}